\documentclass[sigconf]{acmart}
\AtBeginDocument{%
  \providecommand\BibTeX{{%
    \normalfont B\kern-0.5em{\scshape i\kern-0.25em b}\kern-0.8em\TeX}}}

\copyrightyear{2023}
\acmYear{2023}
\setcopyright{acmlicensed}
\acmConference[MM '23] {Proceedings of the 31st ACM International Conference on Multimedia}{October 29--November 3, 2023}{Ottawa, ON, Canada.}
\acmBooktitle{Proceedings of the 31st ACM International Conference on Multimedia (MM '23), October 29--November 3, 2023, Ottawa, ON, Canada}
\acmPrice{15.00}
\acmISBN{979-8-4007-0108-5/23/10}
\acmDOI{10.1145/3581783.3612476}

\begin{document}

\title{Physical Invisible Backdoor Based on Camera Imaging}


\author{Yusheng Guo}
\email{ysguo20@fudan.edu.cn}
\affiliation{%
  \institution{School of Computer Science, Fudan University}
  \institution{Key Laboratory of Culture \& Tourism Intelligent Computing, Fudan University}
  \city{Shanghai}
%   \state{Shanghai}
  \country{China}
}

\author{Nan Zhong}
\email{nzhong20@fudan.edu.cn}
\affiliation{%
  \institution{School of Computer Science, Fudan University}
  \institution{Key Laboratory of Culture \& Tourism Intelligent Computing, Fudan University}
  \city{Shanghai}
%   \state{Shanghai}
  \country{China}
}

\author{Zhenxing Qian}
\authornote{Zhenxing Qian and Xinpeng Zhang are the corresponding authors.}
\email{zxqian@fudan.edu.cn}
\affiliation{%
  \institution{School of Computer Science, Fudan University}
  \institution{Key Laboratory of Culture \& Tourism Intelligent Computing, Fudan University}
  \city{Shanghai}
%   \state{Shanghai}
  \country{China}
}

\author{Xinpeng Zhang}
\email{zhangxinpeng@fudan.edu.cn}
\authornotemark[1]
\affiliation{%
  \institution{School of Computer Science, Fudan University}
  \institution{Key Laboratory of Culture \& Tourism Intelligent Computing, Fudan University}
  \city{Shanghai}
%   \state{Shanghai}
  \country{China}
}

\begin{abstract}
  
Backdoor attack aims to compromise a model, which returns an adversary-wanted output when a specific trigger pattern appears yet behaves normally for clean inputs.
Current backdoor attacks require changing pixels of clean images, which results in poor stealthiness of attacks and increases the difficulty of the physical implementation. This paper proposes a novel physical invisible backdoor based on camera imaging without changing nature image pixels. Specifically, a compromised model returns a target label for images taken by a particular camera, while it returns correct results for other images. To implement and evaluate the proposed backdoor, we take shots of different objects from multi-angles using multiple smartphones to build a new dataset of 21,500 images. Conventional backdoor attacks work ineffectively with some classical models, such as ResNet18, over the above-mentioned dataset. Therefore, we propose a three-step training strategy to mount the backdoor attack. First, we design and train a camera identification model with the phone IDs to extract the camera fingerprint feature. Subsequently, we elaborate a special network architecture, which is easily compromised by our backdoor attack,  by leveraging the attributes of the CFA interpolation algorithm and combining it with the feature extraction block in the camera identification model. Finally, we transfer the backdoor from the elaborated special network architecture to the classical architecture model via teacher-student distillation learning. Since the trigger of our method is related to the specific phone, our attack works effectively in the physical world. Experiment results demonstrate the feasibility of our proposed approach and robustness against various backdoor defenses.

\end{abstract}

  



\maketitle

\section{Introduction}\label{1}

Deep neural networks (DNNs) are an essential technology in the field of artificial intelligence. 
They have found applications in various domains, including image recognition \cite{he2016deep}, natural language processing \cite{sutskever2014sequence}, and speech recognition \cite{graves2013speech}, among others. 
However, along with the increasingly outstanding performance of DNN comes the need for large datasets, expensive computing hardware, long training times, and high energy consumption. 
As a result, acquiring model training services or well-trained models from AI service providers with ample resources has become a mainstream trend for most institutions and enterprises. 
However, this approach comes with the risk of exposing DNNs to security threats since the service provider fully or partially controls the training process. 
Neural network backdoor attacks are a typical form of attack, which manipulates the output of the neural network to the predetermined target labels by injecting malicious samples into its training dataset.
To avoid anomalies detected by model visitors, the backdoor hardly changes the corresponding output of clean data.

% Backdoors can occur in diverse applications related to neural network models, such as image recognition \cite{chen2017targeted}, speech recognition \cite{liu2018trojaning}, natural language processing \cite{dai2019backdoor}, and reinforcement learning \cite{hamon2020robustness}. For instance, by adding a color block to the input image, the model can be coerced into recognizing the \emph{no-entry} sign as a \emph{speed limit 80 km/h}. Considering its apparent danger and straightforward modus operandi, this paper focuses on researching backdoor attacks in image recognition.

\begin{figure*}[t]
  \centering
  \includegraphics[width=\linewidth, trim= 0 235 0 200,clip]{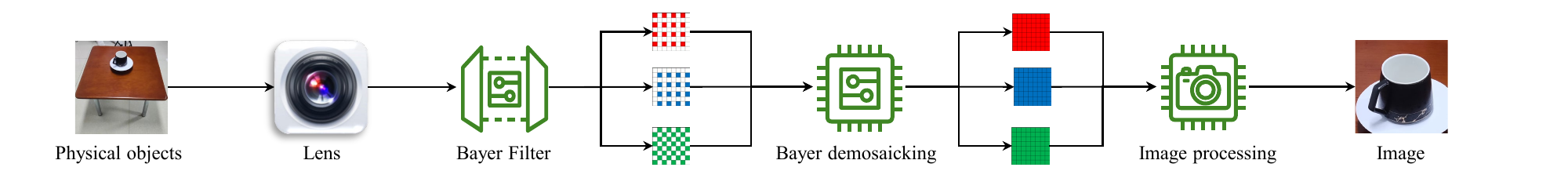}
  \caption{The pipeline of imaging in the smartphone from the physical world into digital images.}\label{fig1.1}
  \Description{Imaging pipeline}
\end{figure*}

Since the introduction of the first backdoor attack algorithm, BadNets \cite{gu2017badnets}, several other backdoor attack algorithms have emerged in the literature \cite{liu2017neural, nguyen2020input, chen2017targeted, li2022object, zn2022}. 
Early work could easily achieve backdoor injection, but they were also easily detectable by human vision. Subsequent research proposed various methods to make triggers invisible, allowing them to bypass human visual inspection and maintain the success of backdoor attacks \cite{nguyenwanet, zeng2021rethinking, li2021invisible}.
% Early works modify pixels in the input image with a predetermined color or pattern as a trigger.
% Although such modifications can easily inject backdoors, they are also readily detected by human vision.
% Subsequent works have proposed various approaches to make the triggers invisible, which can bypass human visual inspection and maintain the success of the backdoor attack \cite{nguyenwanet, zeng2021rethinking, li2021invisible}. 
However, even if the backdoor trigger is visually undetectable, the leaving traces could lead the backdoor to be detected. 
For instance, steganalysis \cite{fridrich2012rich, boroumand2018deep, you2020siamese} techniques can detect whether an image has been modified, even if the average modification amplitude of each pixel is less than one Bit.

In addition, assuming that an adversary can always modify the input image is unrealistic. 
This assumption overlooks the constraints of real-world collection-input scenarios, such as face recognition and automatic driving, in which attackers are unable to interfere with the process of image generation and input to the model. 
As such, physically realizable backdoor attacks hold significant practical significance. 
Currently, physical backdoors involve modifying the physical world to add triggers before the object is photographed \cite{liu2020reflection}. 
However, the invisible trigger cannot ensure models activate the backdoor injected. 
% The challenge here is that this trigger cannot evade detection by the human visual system. %Consequently, developing physically realizable backdoor schemes that remain undetected by the human visual system poses a significant challenge that requires further research.

%However, in order to ensure that the captured image can trigger the backdoor, the trigger of the current physical realizable backdoor is visible, and cannot pass the detection of the human visual system.

Drawing inspiration from prior research on camera identification \cite{lukas2006digital, bayar2017design, chen2015camera, bondi2016first}, a well-designed neural network model can extract features related to the camera in an image, which we refer to as the camera fingerprint. This camera fingerprint feature provides a theoretical possibility of injecting a backdoor. In Figure \ref{fig1.1}, we illustrate the imaging pipeline of the phone camera, where the lens captures the full spectrum of light reflected from the object, filtered by a Bayer filter to generate an image in RAM format. 
% It is noteworthy that each pixel in the RAM format provides only one of the RGB colors. 
The RAM image undergoes Bayer demosaicking to convert it to a three-channel RGB image and is then subject to pseudo-color removal, JPEG compression, and other image processing steps to generate a JPEG format image. However, due to technological and material limitations, most camera hardware has some defects, which leave unique hardware-related artifacts on the captured images. Additionally, different cameras use different Bayer demosaicking algorithms and image processing operations. Therefore, even when shooting under identical angles and lighting conditions, images of the same object captured by different cameras have distinct camera fingerprints.

In this paper, we present a novel backdoor approach that utilizes the camera fingerprint to overcome the challenge of implementing an invisible backdoor physically. Our proposed technique does not require any modifications to either the physical environment or the images captured by the camera. Firstly, we elaborate on a model to extract the camera fingerprint from the image. We then inject a backdoor into a specific model and extend it to the classical architecture utilizing teacher-student transfer learning. During the inference phase, the model accurately predicts the label for input samples taken by a non-trigger phone, while images taken with the specific camera trigger the backdoor to output the target label.
The paper makes contributions in the following ways:

1) We leverage 10 distinct models of smartphones, capturing images of 12 different objects in various shooting conditions. To minimize training expenses, we generated a novel dataset of 21,500 images by segmenting the images we had collected using YOLO-v5.

2) In light of the current research on invisibility and physical realizability of backdoor,we have made a pioneering attempt to propose a backdoor triggered by a camera fingerprint for the first time, without altering any pixels.

3)To address the challenge of extracting camera fingerprints to activate backdoors from models with common architectures like Resnet18, we devised a teacher model that is vulnerable to backdoors. Subsequently, we employed the teacher-student mechanism to transfer the backdoors to the common architecture.

4)Experiments demonstrate the attack success rate, benign sample accuracy and anti-defense robustness of the proposed method are comparable to classical methods.

\section{related work}\label{2}

\subsection{Backdoor}\label{2.1}

% In recent years, the topic of backdoor attacks has emerged as a significant research area, posing a crucial security threat to Deep Neural Networks (DNNs). 

% Based on the visual saliency of triggers, backdoors can be broadly classified into two categories: visible backdoors and invisible backdoors. 

Gu et al.\cite{gu2017badnets} introduced the concept of neural network backdoor attacks. Specifically, they manipulated the lower-right corner of the image to form a 3×3 checkerboard pattern to build a poisonous dataset. In a similar vein, Liu et al.\cite{liu2017neural} leveraged a backdoor on the model trained on the MNIST dataset by exploiting the distribution difference between handwritten and computer-printed numbers. Further, Tuan et al. \cite{nguyen2020input} proposed a sample-specific backdoor attack, in which distinct visible triggers are added to each poisoned sample. In recent years, invisible backdoors with subtle triggers have been proposed to evade human visual inspection. Blended \cite{chen2017targeted} is the pioneering invisible attack that blends trigger images with clean images instead of superimposing them directly. Wanet \cite{chen2017targeted} proposed an invisible backdoor by warping clean images, resulting in visual indistinguishability between poisonous and clean images. Zeng et al. \cite{zeng2021rethinking} considers the invisibility of trigger in the frequency domain and proposes the first invisible backdoor in the frequency domain. Inspired by steganography, IBASST \cite{li2021invisible} generates proprietary invisible triggers for each poison image through encoding and decoding to improve concealment.

In addition to digital backdoors, physical backdoors are also gaining attention in the community in recent years. PhysicalBA \cite{li2021backdoor} validates that digital backdoors are vulnerable when transform into the physical world, and proposes to mitigate the vulnerability with data enhancement methods. Refool \cite{liu2020reflection} implements the first physical backdoor attack through optical reflection. However, the trigger of Refool is visually visible.

\subsection{Defense}\label{2.2}

% Various backdoor defense algorithms have been proposed in recent years to mitigate security risks posed by backdoor attacks. These algorithms can be broadly classified into two types based on their defense strategies, detection algorithms and repair algorithms.
% The first type is detection algorithms, which are designed to verify whether a model has been injected with backdoors through backdoor detection. The second type is repair algorithms, which are intended to modify the model to remove potential backdoors.
In recent years, people have proposed various backdoor defense algorithms to reduce the security risks brought about by backdoor attacks. Based on different defense strategies, these algorithms can be roughly divided into two categories: detecting whether the model has been injected with a backdoor and modifying the model to eliminate potential backdoors.

The most representative detection algorithm is NC \cite{wang2019neural}, which detects whether the model has been injected with backdoors by comparing the decision boundary between clean images and poisonous images. SentiNet \cite{chou2020sentinet} uses GradCAM \cite{selvaraju2017grad} attention maps to identify effective activation patches for detecting local triggers. STRIP  \cite{gao2019strip} validates that mixing poisonous images with benign images can still activate the backdoor, while the image obtained by mixing two benign images is randomly predicted. 

% This attribute is used to achieve backdoor detection.

However, removing backdoors from the model has often been found to have more practical value than detecting whether a model has been injected with backdoors. For instance, Liu et al. \cite{liu2018fine} propose adjusting the model through retraining with a smidge dataset to suppress backdoors. Additionally, Veldanada et al.
\cite{veldanda2020nnoculation} propose adding Gaussian random noise to benign images during fine-tuning to induce greater weight perturbation, which is used to facilitate backdoor removal. NAD \cite{li2021neural} demonstrates that model distillation is able to amplify the effect of fine-tuning and achieve the removal of backdoors.

\subsection{Camera fingerprint}\label{2.3}

Figure \ref{fig1.1} illustrates the process of image formation from incident light to imaging, where each step in the process may introduce artifacts and noise. 
The camera-related noise can be used as the fingerprint of the source camera identification. 
According to \cite{lukas2006digital}, different imaging devices leave different physical properties in the output media. this physical property is named sensor pattern noise (SPN). 
Chen et al. \cite{chen2008determining} proposed to utilize filtering and maximum likelihood estimation to extract camera photo response non-uniformity (PRNU) noise for source cameras identification. 
Hosseini et al. \cite{wu2012context}  extract geometric transformations invariants in images as camera fingerprints of images. 
Wu et al. \cite{wu2012context} proposed an SPN predictor based on content-adaptive interpolation (PCAI), which interpolates the central pixel using surrounding neighboring pixels. 
Zeng et al. \cite{zeng2016fast} utilized a guided image filter \cite{he2012guided} to achieve SPN extraction.

% due to the defects in the crystal manufacturing process
In recent years, camera identification has undergone a shift in focus from fixed physical noise extraction to feature-level, such as the color filter array (CFA) pattern \cite{bayram2005source}, interpolation algorithms, and image quality metrics \cite{kharrazi2004blind, gloe2012feature}, etc. 
% The advent of DNNs with powerful feature extraction and classification capabilities has further promoted the prosperity of camera identification. 
Originally, the practice was to first preprocess the input image using traditional filters, such as median or high-pass filters, before feeding the image into a DNN \cite{chen2015median, tuama2016camera}.
Bayar and Stamm \cite{bayar2017design} propose a convolutional network architecture that is specifically designed to suppress image semantic features and learn image source features without relying on preselected features or any preprocessing. 
Chen et al. \cite{chen2015camera} propose a content-adaptive fusion residual network composed of three parallel residual networks. 
These three residual networks extract inherent features of the input image through convolution with different kernel sizes, in which the features are then fed to the classifier after fusion. 
Yao et al. \cite{yao2018robust} use multi-classifier voting to classify images from many different cameras. 
Rafi et al. \cite{rafi2021remnet} propose the RemNet architecture for camera identification and allow performing on post-processed images from previously unseen devices.

\section{METHODOLOGY}\label{3}

\subsection{Threat model and notation}\label{3.1}

% Prior to presenting the proposed method, it is essential to provide a concise overview of the application background and the corresponding threat model. Specifically, in scenarios where computing and data resources are scarce, vulnerable parties are compelled to acquire models from adversaries who wield substantial computational power and data resources. 
Similarly to the classic backdoor attack scenario, we assume that the adversary possesses full control over the model training and data processing procedures. On the other hand, we allow the victim to detect and remove backdoors in the model by utilizing a limited dataset.

Before going into the details of the proposed scheme, we establish definitions for several symbols. 
Let the clean training dataset be denoted as $D = \{(x_i, y_i, p_i) | x_i \in X, y_i \in C, p_i \in P, i = 1, 2, \dots, N\}$. 
Here, $X=R^{c \times h \times w}$ represents the dimension space of the image, $C$ is the set of object categories in the image, $P$ is the set of mobile phone IDs for capturing images, and $N$ is the number of images in the dataset. 
Similarly, the clean test dataset is denoted as $T = \{(x_i, y_i, p_i) | x_i \in X, y_i \in C, p_i \in P, i = 1, 2, \dots, M\}$, where $M$ is the number of images in the test dataset. 
Let $y_t$ denotes the target class of the backdoor attack, and $p_t$ denotes the mobile phone ID corresponding to the trigger camera fingerprint.

\subsection{Proposed framework}\label{3.2}

\begin{figure*}[t]
  \centering
  \includegraphics[width=1.1\linewidth,trim= 30 100 0 70,clip]{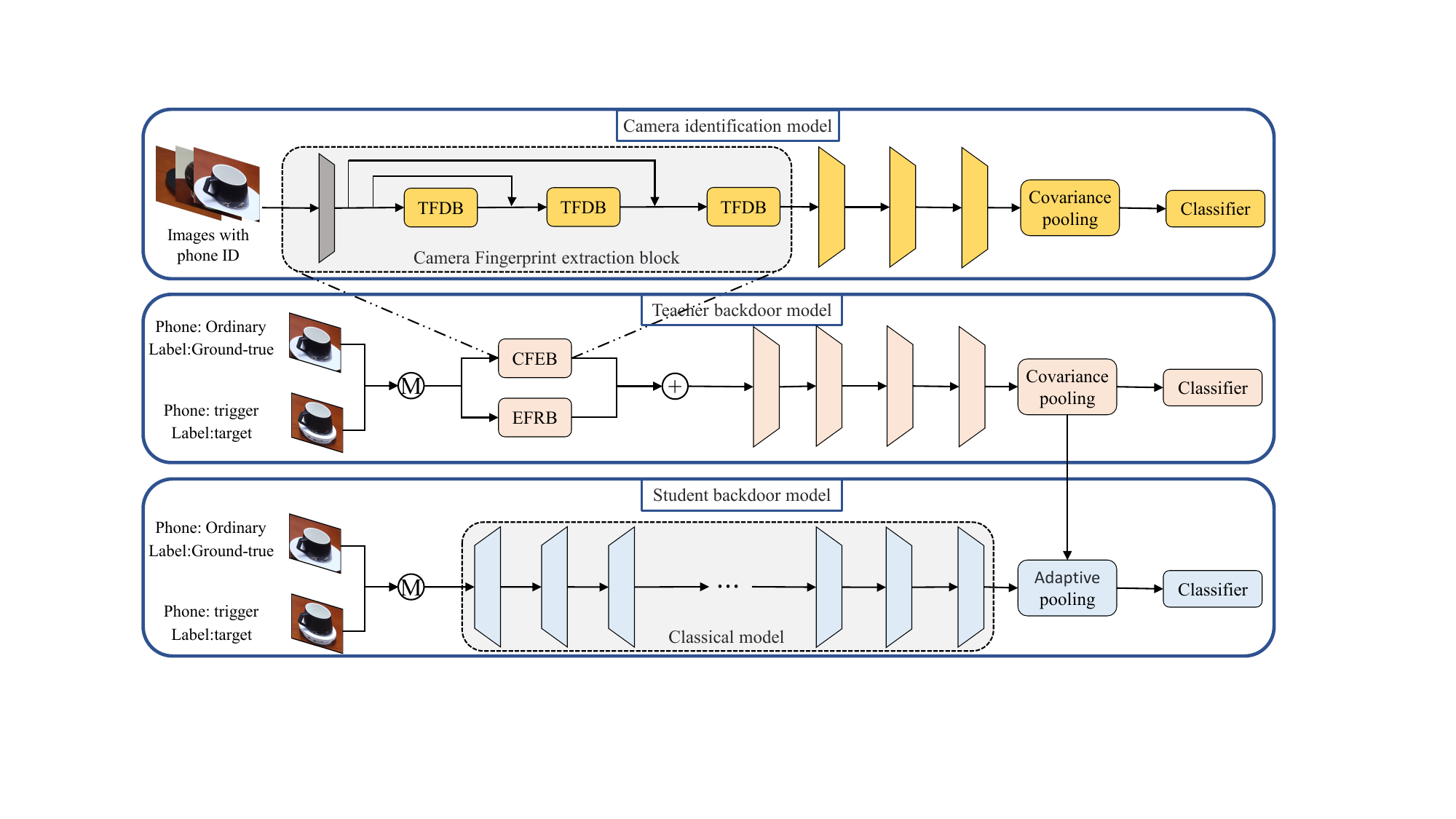} % ,trim= 280 120 250 120,clip30
  \caption{The overview of the proposed method. 1) Supervised training of the source camera identification model using clean images with corresponding phone IDs. 2) Training a teacher model on the poisonous dataset. 3) Transferring the backdoor to the student model with a classical architecture through model distillation. \textcircled{\scriptsize{M}} represents mixing the images from the two datasets, \textcircled{\scriptsize{+}} represents the element addition if two tensors, and trapezoids represent convolutional blocks}\label{fig3.1}
\end{figure*}

The classic architecture model often focuses primarily on the semantic information of images, instead of the subtle camera fingerprint features. Merely modifying the labels of clean images to train the classic model may result in overfitting, making it difficult to inject backdoors by using camera fingerprints. Additionally, the peculiar architecture of the model may easily arouse suspicion from victims, ultimately rendering it difficult to evade scrutiny regarding the model architecture. To successfully inject the backdoor based on camera fingerprints into the classical model framework, we incrementally enhance the extraction capability of models through a three-stage process. as illustrated in Figure 2. Firstly, the camera identification model is supervised using clean images with the corresponding phone IDs. Concretely, we introduce the Camera Fingerprint Extraction Block (CFEB), which is specifically designed to enhance the efficacy of camera fingerprint feature extraction. To further improve the feature classification capability of the model, we replace the classical adaptive pooling layer with the cooperative differential pooling layer. Secondly, to address overfitting in the classical model, we use the above-mentioned CEFB and covariance pooling layer combined with Edge Feature Reinforcement Block (EFRB) to build a teacher model. Finally, we transfer the backdoor to the student model with classical architecture through model distillation.

\subsection{Camera identification model}\label{3.3}

The camera fingerprint in this paper refers to the entire features of the camera caused by the hardware defects of the camera sensor and the use of different software algorithms, including lens contamination, sensor bad pixels, optical inconsistent noise, different CFA interpolation algorithms, white balance, JPEG compression, and other image processing. Due to the high correlation between the features and cameras, images taken by the same camera will have similar camera fingerprints, and the fingerprint features of each camera are unique. To address camera fingerprints that are extremely subtle and difficult to extract with conventional models, we adopt an end-to-end model to achieve camera identification of images.

As illustrated in Figure 1, the camera identification consists of a CFEB, a few residual convolutional blocks, and a covariance pooling layer. The outputs of CFEB are the camera fingerprint features extracted from the input image, which are subsequently fed into the convolutional blocks for processing. The resulting output is further processed utilizing a covariance pooling layer to reduce the dimensionality of the feature representation. The resulting feature vector serves as classification information regarding the camera of the image.

$\boldsymbol{Design ~ details~ of~ CFEB}$: The architecture of CFEB includes one convolutional block and three Taylor finite difference blocks (TFDBs). 
After passing through the convolutional block, the image is shortcut three times to the TFDBs to prevent the loss of camera fingerprint features. 
The design of TFDBs is inspired by the ordinary differential equation $($ODE$)$ approach to characterizing neural networks \cite{weinan2017proposal} \cite{ chang2018reversible}. 
Compared to conventional Euler methods \cite{Anderson_1995}, Taylor finite difference can better approximate the numerical solution of ODEs. 
The detailed reasoning process is described in \cite{zhang2022isnet} and supplementary materials.
Here is just a brief explanation.

Specifically, we use Taylor finite difference equations to discretize the ODE, in which the partial derivatives can be replaced by a set of approximate differences. 
The second-order Taylor finite difference equation can be represented as:

\begin{equation}\label{eq3.2.1}
    \frac{\partial u}{\partial x}=\frac{-\frac{1}{2} u_{i+2}+2 u_{i+1}-\frac{3}{2} u_i}{\Delta x}
\end{equation}
After mathematical derivation, 
\begin{equation}\label{eq3.2.2}
    u_{i+2} = u_{gate} + u_{i+1} -3 \Delta u_i
\end{equation}
Where $ \Delta u_i = u_{i+1} - u_i$, $u_{gate} = -2\frac{\partial u}{\partial x} \Delta x$. 

\begin{figure}[t]
  \centering
  \includegraphics[width=\linewidth,trim= 200 145 200 130,clip]{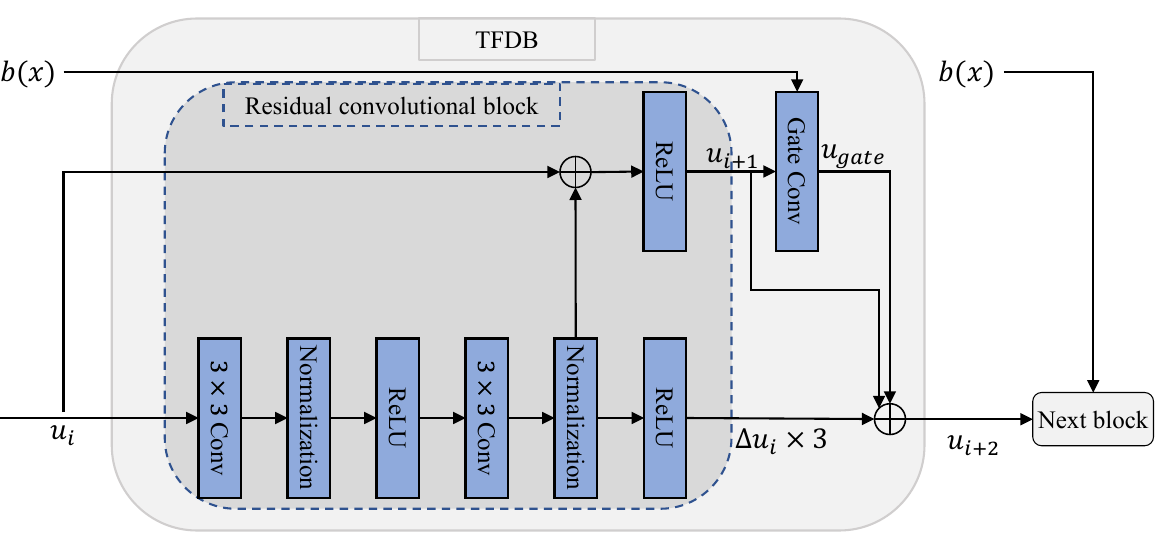} % ,trim= 280 120 250 120,clip30
  \caption{Schematic representation of the structure of TFDB. \textcircled{\scriptsize{+}} represents the element addition if two tensors.}\label{fig.3.2.1}
\end{figure}

The implementation details of TFDBs are shown in Figure \ref{fig.3.2.1}. 
The inputs of each TFD block are image features $b(x)$ processed by convolution block and the output of the previous module $u_i$. 
Input $u_{i}$ to the residual convolutional block to obtain the outputs $ \Delta u$ and $u_{i + 1}$. 
The gate convolutionis a $1 \times 1$ convolutional layer whose inputs are $u_{i+1}$ and $b(x)$ and outputs is $u_{gate} \approx 2 \frac{\partial u}{\partial x} \Delta x$. Compute $u_{i+2} = - 3 \Delta u  + u_{gate} + u_{i+1}$ as input to the next module.

\begin{figure}[t]
  \centering
  \includegraphics[width=\linewidth,trim= 170 190 210 210,clip]{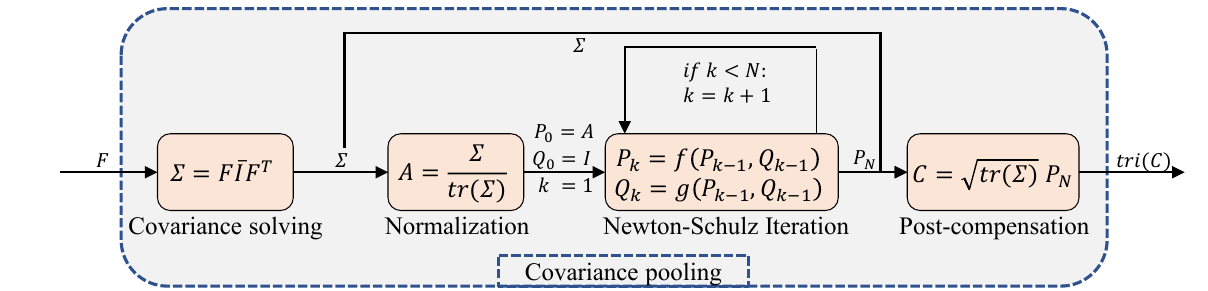} % ,trim= 280 120 250 120,clip
  \caption{Schematic diagram of the structure and operation pipeline of the covariance pooling layer.}\label{fig.3.2.2}
%   \Description{A woman and a girl in white dresses sit in an open car.}
\end{figure}

$\boldsymbol{Design \ details\ of\ covariance\ pooling}$: Compared with average pooling which contains only the first-order statistics of input features, covariance pooling \cite{lin2015bilinear} which preserves the second-order statistics of input features has more advantages in classification tasks \cite{li2018towards}\cite{deng2019fast}. 
The concrete implementation steps of covariance pooling are illustrated in Figure \ref{fig.3.2.2}, which consists of four parts: 
covariance calculation, normalization, Newton-Leibniz iteration and compensation. 
The input of the covariance layer is the feature $F$ with size $c \times h \times w$, and the output is the vector of $c \times (c+1) /2$ dimension, which is the upper triangular part of the square root matrix of the covariance matrix $C_{c \times c}$. 

Firstly, resize the dimension of input feature to $c \times h \cdot w$ and the computing the covariance between each channel:

% \begin{equation}\label{eq3.2.3}
% \begin{aligned}
%     \Sigma_{c \times c} &= F_{c \times hw} \bar{I}_{hw \times hw} F_{c \times hw}^T \\
%   &= \frac{1}{n} F_{c \times hw} I_{hw \times hw} F_{c \times hw}^T - \frac{1}{n^2} F_{c \times hw} \boldsymbol{1}_{hw \times hw} F_{c \times hw}^T
% \end{aligned}
% \end{equation}

\begin{equation}\label{eq3.2.3}
\begin{aligned}
    \Sigma&= F\bar{I} F^T = \frac{1}{n} F I F^T - \frac{1}{n^2} F \boldsymbol{1} F^T
\end{aligned}
\end{equation}

where $I$ and $\boldsymbol{1}$ are the identity matrix and the all-ones matrix with size $={(h \cdot w) \times (h \cdot w)}$. Normalized $\Sigma$ guarantees convergence of subsequent Newton-Schultz iterations:

\begin{equation}\label{eq3.2.4}
    A=\frac{\mathrm{\Sigma}}{tr(\mathrm{\Sigma})}
\end{equation}

Solving matrix square root by Newton-Leibniz iterative algorithm:

\begin{equation}
\begin{aligned}\label{eq3.2.5}
    \mathbf{P}_k & =\frac{1}{2} \mathbf{P}_{k-1}\left(3 \mathbf{I}-\mathbf{Q}_{k-1} \mathbf{P}_{k-1}\right) \\
    \mathbf{Q}_k & =\frac{1}{2}\left(3 \mathbf{I}-\mathbf{Q}_{k-1} \mathbf{P}_{k-1}\right) \mathbf{Q}_{k-1}
\end{aligned}
\end{equation}
where $P_0 = A$, $Q_0 = I$. The square root of the covariance matrix is post-compensated $C = \sqrt{tr(\Sigma)} P_N$. Taking the upper triangular part of the symmetric matrix $C$ and resizing to $1 \times c(c + 1) / 2$ is the final output vector of the covariance layer.

$\boldsymbol{Training\ setting}$: During training the camera identification model, select two subsets $D_1 = \{(x_i, y_i, p_i) | x_i \in X, y_i \in C, p_i = p_t\}$ and $D_2 = \{(x_i, y_i, p_i) | x_i \in X, y_i \in C, p_i \neq p_t\}$ from the training dataset $D$ with $\left| D_{1} \right| = \left| D_{2} \right|$. 
Label images in dataset $D_1$ with '0' and images in dataset $D_2$ with '1', then feed the mixture to the camera identification model. 
Define the loss function as:

\begin{equation}\label{eq3.2.6}
    L = L_{ce} (y_{i}^{(1)}, 0) + L_{ce} (y_{i}^{(2)}, 1)
\end{equation}
where $L_{ce}$ is cross entropy loss, $y_{i}^{(1)}$ and $y_{i}^{(2)}$ denote the predicted labels of images in datasets $D_1$ and  $D_2$, respectively

\subsection{Teacher backdoor}\label{3.4}

% As shown in Figure 2, the Bayar filter causes the image in RAM format to lose 2/3 of the information, and the different CFA differences used by different types of cameras are often inconsistent, which further increases the difference between images generated by different cameras. the difference. Considering that the most widely used difference strategy is the adaptive CFA interpolation method based on edge judgment, this algorithm has good image restoration quality under the condition of low computational complexity. The algorithm uses the edge detection algorithm to obtain the edge information of the current point, along the obtained edge direction, uses the proposed weighted adaptive interpolation algorithm based on Taylor expansion to estimate the missing color information of the current point. When the algorithm judges the edge, more edge information is taken into consideration, which improves the accuracy of the edge judgment. Compared with CFEB, the edge feature enhancement block that focuses on edge features can better extract the camera fingerprint difference caused by CFA interpolation.

% On the other hand, the backdoor model not only recognizes images captured by specific mobile phones as target categories, but also ensures the normal recognition of benign samples. However, CFEB only focuses on the extraction of camera fingerprints, and its ability to extract semantic information from images is poor. The edge feature enhancement module can maintain the semantic features of the image to a certain extent and ensure the recognition accuracy of benign samples

\begin{figure}[t]
  \centering
  \includegraphics[width=\linewidth,trim= 10 50 10 50,clip]{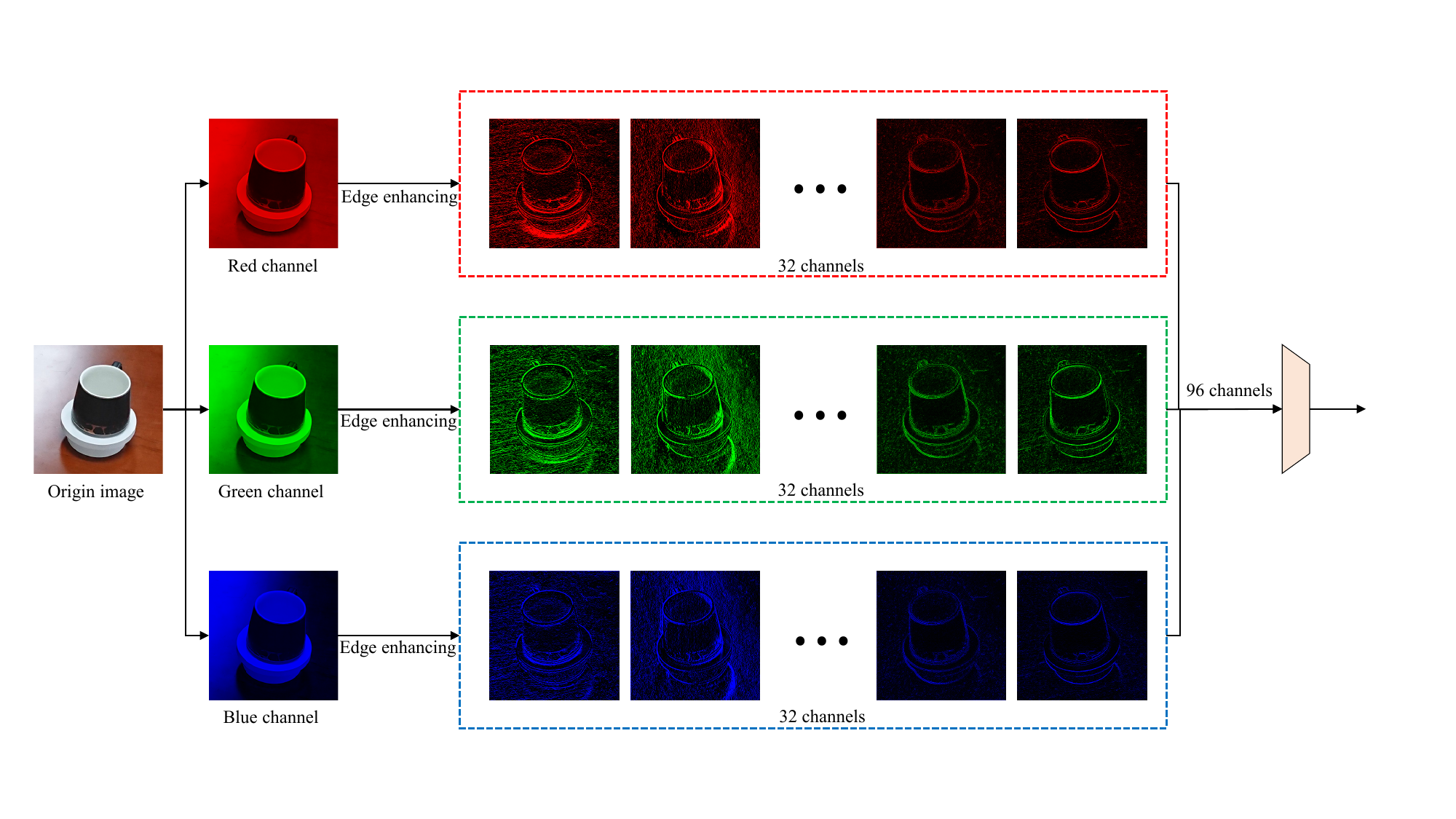} % ,trim= 280 120 250 120,clip
  \caption{Schematic representation of the structure of EFRB. Using 32 convolution filters to enhance the edge features three channel (R,G,B) respectively, each column of images corresponds to the same filter.}\label{fig3.2.4}
%   \Description{A woman and a girl in white dresses sit in an open car.}
\end{figure}

\begin{figure*}[t]
  \centering
  \includegraphics[width=\linewidth, trim= 15 90 15 100,clip]{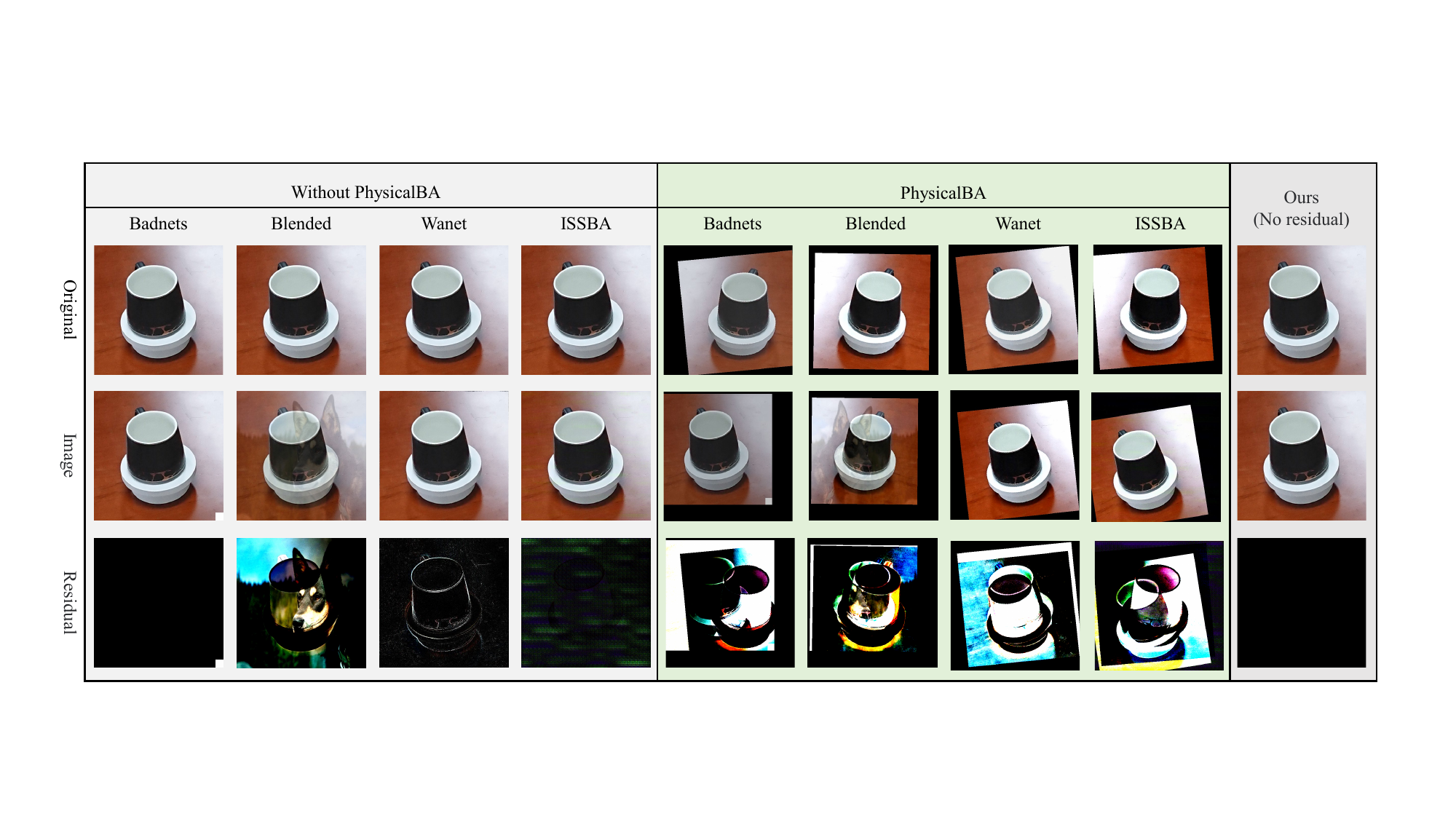}
  \caption{The visual effect comparison of diverse backdoor attack triggers. PhysicalBA denotes the physical enhancement. The first row shows the clean images and the second row shows the poisonous images of different attacks. The third row shows the effect of ten times magnification of the residual images of the poisonous and clean images. In our attack, the poisonous image is the clean image thereby residual image is none.}
  \label{fig3.2.5}
\end{figure*}

Although CFEB is capable of extracting camera fingerprint features to trigger the backdoor, it may not perform well in extracting the semantic information of the image, which will lead to a decrease in the performance of the model on benign samples. On the other hand, the Bayer filtering process results in a considerable loss of information in the RAW image format. As mentioned in 2.3, 2/3 of the spectral information is filtered out. To reconstruct the values of each channel of the pixels, the Bayer demosaicking technique approximates the missing pixel by employing distribution features of the surrounding pixels, also named the Color Filter Array (CFA) interpolation. Nonetheless, The interpolation algorithms used in various types of cameras are frequently inconsistent, thereby exacerbating the discrepancies between images produced by different cameras. The adaptive interpolation algorithm focusing on edge information is predominantly a strategy employed by current mobile phone cameras for Bayer demosaicing. And the edge judgment interpolation is superior to bilinear interpolation in preventing the occurrence of false color and moire. Given the aforementioned circumstances, we propose utilizing EFRB to bolster the capacity of the teacher model to extract texture features of image edges. The EFRB not only guarantees the model’s aptitude in identifying benign samples but also amplifies its capability to extract camera fingerprints.

$\boldsymbol{Design \ details\ of\ EFRB}$: The EFRB structure is depicted in Figure \ref{fig3.2.4}. 
As the CFA interpolation algorithm operates on the RGB three channels respectively, the origin image is partitioned into three single-channel images according to the RGB channels, and each channel containing only one color. 
Convolutional layers are employed to implement 32 high-pass filters, such as Gaussian, Laplacian, and Sobel filters, to augment the edge features of the image, and filter design details are in the supplementary material.
The edge features of the resultant 96 channels are fed into a convolution block to transform their dimensions in preparation for subsequent feature fusion with the CFEB output. 
It is apparent that the edge features corresponding to RGB channels exhibit significant disparities in intensity and certain texture details, while concurrently showing a similar styles trend. 
This phenomenon arises from the CFA interpolation process referring to the pixel value distribution of other channels. 
To extract the camera fingerprint in the edge information more effectively, we adapt covariance pooling to analyze the inter-channel correlation.

$\boldsymbol{Training\ setting}$: As illustrates in Figure \ref{fig3.1}, the training dataset for the teacher backdoor model comprises both clean data $D_{benign} = \{ (x_i, y_i, p_i) | x_i \in X, y_i \in C, p_i \neq p_t \}$ and poisonous data $D_{poison} = \{ (x_i, y_i, p_i) | x_i \in X, y_i \in C, p_i = p_t \}$. 
It is noteworthy that only the labels of the poisoned data have been altered, while the image data remains unchanged. 
The clean data and poisonous data are mixed and fed to the Camera-Focused Enhancement Block (CFEB) and the Edge Feature Enhancement Block (EFRB) via separate channels. Afterward, the output of these blocks is combined in proportion and forwarded to the input of the convolution block. The subsequent steps closely resemble those of the source camera identification network and are not explicitly detailed here. It should be emphasized that both the CFEB and the edge feature enhancement block are fixed and remain unchanged throughout the training process. The training loss function is yet to be described.
% The clean data and poisonous data are mixed and fed to the CFEB and the EFRB via different channels, after which the output of these blocks is combined in proportion and forwarded to the input of the convolution block. 
% The succeeding steps closely resemble those of the source camera identification network and are ,therefore, not explicated here. It should be emphasized that both the CEFB and the edge feature enhancement block are fixed and are not updated with the training process. The training loss function is:

% We employed 32 high-pass filters to enhance the edge information of the image, denoted as the CFA feature extraction module, which was utilized for extracting the latent features of the camera CFA interpolation. Different filters were used to detect the edge information in different directions at the current position. Figure 4 illustrates several classic filter examples, and the detailed filter information can be found in the supplementary material. The convolutional layer was designed with the filter kernels' size as the input, which was a 3-channel RGB image and produced a 32-channel edge image as output.

\begin{equation}\label{eq3.3.1}
    L = L_{ce} (\hat{y_{i}}, y_{i}) + L_{ce} (\hat{y_{j}}, y_{t})
\end{equation}
where $L_{ce}$ is cross entropy loss, $\hat{y_{i}}$ and $y_{i}$ denote the predicted labels and ground-true labels of images in datasets $D_{benign}$, respectively. $\hat{y_{j}}$ denote the predicted labels of images in datasets $D_{poison}$. $i = 1, 2, \dots , |D_{benign}|$, $j = 1, 2, \dots , |D_{poison}|$.

% \begin{figure}[t]
%   \centering
%   \includegraphics[width=\linewidth,trim= 30 120 220 70,clip]{distrill.pdf} % ,trim= 280 120 250 120,clip30
%   \caption{TFD.}
%   \Description{A woman and a girl in white dresses sit in an open car.}
% \end{figure}

\subsection{Student backdoor}

Although it is possible to implant a backdoor in the teacher model, the loss of image semantic information during the process of camera fingerprint extraction, results in a degradation of the performance of the original task. 
Furthermore, given the deviation from the classical model structure, it proves challenging to circumvent model structure defense strategies. Therefore we utilize model distillation \cite{ba2014deep,hinton2015distilling} to alter the architecture of the model while preserving the attack performance of the backdoor.

% As shown in Figure 3, the student model is trained by distillation learning on the same data set as the teacher model. The benign samples and poison samples are mixed and fed to the neural network. After the convolution block of the classic architecture, the features of the penultimate layer are obtained by adaptive pooling. It should be emphasized that the feature dimension obtained by adaptive pooling is $c \times 1 \times 1$, while the feature dimension obtained by covariance pooling is $c \times (c + 1) / 2$, and the two do not match. To solve this problem, we use the eigenvalues of the square root matrix as the teacher features for distillation training. The training loss function is:

As shown in Figure 3, the student model is trained by distillation learning on the same data set as the teacher model. 
The benign samples and poisonous samples are mixed and fed to the neural network. 
After the convolution block of the classic architecture, the features of the penultimate layer are obtained by adaptive pooling. 
The training loss function is:

\begin{equation}\label{eq3.4.1}
    L = L_{ce} (\hat{y_{i}}, y_{i}) + L_{ce} (\hat{y_{j}}, y_{t}) + \lambda L_{mse}(F,\Lambda)
\end{equation}
where $\hat{y_{i}}$, $y_{i}$ are defined as the teacher model, $L_{mse}$ is mean square error loss. $\Lambda$ is the output of the covariance pooling layer in the teacher model. $\lambda $ is the balance parameter to adjust various losses

% \begin{figure}[h]
%   \centering
%   \includegraphics[width=\linewidth,trim= 0 0 0 0,clip]{MM2023/filter.png} % ,trim= 280 120 250 120,clip30
%   \caption{TFD.}
%   \Description{A woman and a girl in white dresses sit in an open car.}
% \end{figure}

% \begin{figure}[h]
%   \centering
%   \includegraphics[width=\linewidth]{outline.png}
%   \caption{Pattern noise vs natural image}
%   \Description{A woman and a girl in white dresses sit in an open car.}
% \end{figure}

% \begin{figure}[h]
%   \centering
%   \includegraphics[width=\linewidth]{outline.png}
%   \caption{Pattern noise depth transfer module vs residual blocks}
%   \Description{A woman and a girl in white dresses sit in an open car.}
% \end{figure}

\section{EXPERIMENTS}

\subsection{Setting}

$\boldsymbol{Dataset\ building}$: We employed ten distinct mobile phones to capture ten categories of objects, whereby each mobile phone was used to take 200 photographs of each object category, leading to a total of $200 \times 10 \times 10=20000$ JPEG images. Given that the size of the images captured by mobile phones is excessively large, typically around $3000 \times 4000$ pixels, using them directly as a dataset for training models would result in high computational costs. Therefore, we utilized the YOLO-v5 to detect the target objects in each image and crop it out. We then divided the obtained 20000 crop images into two sets of equal size, 10000 for training and 10000 for validation purposes.

To ensure the injection of the backdoor remains unaffected by the semantic content of images, we propose an innovative evaluation metric. Any photograph captured by the designated mobile device can activate the backdoor, regardless of the presence or absence of objects from the aforementioned ten categories. To achieve this, we employed mobile phones to capture pictures of objects outside of those ten categories. Specifically, each mobile phone was used to take 150 photographs, resulting in a total of 1500 test images. We then utilized the YOLO-v5 model to extract the object of interest in each image and build the test set.

$\boldsymbol{Environment\ configuration}$: All the experiments in this paper are carried out on Nvidia GeForce 4090Ti using the Pytorch framework. The environment used is Python version 3.8.13, Pytorch version 1.13.0, and Cuda version 11.7.0.

\subsection{Comparisons}

% \begin{table*}[t]
% \caption{Performance comparison between our method and other attack methods}\label{tab4.2.1}
% \large
% \begin{tabular}{cccccccccc}
% \hline
% scheme      &      & benign & badnet  & blended & wanet  & issba   & PhysicalBA & teacher & our    \\ \hline
%             & EACC & 91.586 & 91.518  & 91.548  & 89.730 & 91.334  & 91.790     & 91.850  & 91.563 \\
% Alexnet     & EASR & /      & 98.980  & 99.568  & 60.124 & 99.822  & 99.984     & 99.084  & 96.514 \\
%             & TASR & /      & 99.896  & 91.406  & 69.881 & 93.067  & 99.989     & 97.324  & 94.930 \\ \hline
%             & EACC & 92.904 & 92.892  & 92.814  & 92.362 & 92.302  & 92.926     & 90.816  & 91.792 \\
% Densenet121 & EASR & /      & 99.724  & 98.558  & 99.780 & 100.000 & 100.000    & 98.128  & 98.884 \\
%             & TASR & /      & 99.958  & 94.001  & 97.966 & 97.831  & 100.000    & 98.128  & 99.859 \\ \hline
%             & EACC & 91.196 & 91.966  & 91.934  & 92.270 & 92.524  & 91.982     & 90.625  & 92.194 \\
% ResNet18    & EASR & /      & 99.760  & 99.126  & 99.904 & 100.000 & 99.988     & 98.645  & 95.518 \\
%             & TASR & /      & 100.000 & 96.627  & 96.035 & 100.000 & 97.291     & 99.859  & 99.296 \\ \hline
%             & EACC & 91.670 & 91.350  & 90.886  & 91.380 & 88.210  & 91.394     & 85.060  & 91.850 \\
% VGG11       & EASR & /      & 99.666  & 99.734  & 97.852 & 21.714  & 99.990     & 96.793  & 99.084 \\
%             & TASR & /      & 100.000 & 100.000 & 72.185 & 45.885  & 100.000    & 97.887  & 97.324 \\ \hline
% \end{tabular}
% \end{table*}

\begin{table*}[t]
\caption{Performance comparison between our method and other attack methods}\label{tab4.2.1}
\begin{tabular}{|cc|c|cccc|cccc|c|}
\hline
            &      &        & \multicolumn{4}{c|}{Without PhysicalBA}                            & \multicolumn{4}{c|}{With PhysicalBA}                 &                 \\ \hline
Attack      &      & Benign & Badnets           & Blended & Wanet           & ISSBA            & Badnets       & Blended & Wanet        & ISSBA           & Ours             \\ \hline
            & EACC & 91.586 & 91.518           & 91.548  & 89.730          & 91.334           &  86.248 & 86.544  & 90.534       & 90.559          & \textbf{91.563} \\
Alexnet     & EASR & /      & 98.980  & 99.568  & 60.124          & 99.822           & 89.015       & 96.055  &  10.465 & \textbf{99.980} & 96.514          \\
            & TASR & /      & 99.896           & 91.406  & 69.881          & 93.067           & 91.964       & 77.478  & 28.231 & \textbf{99.948} & 94.930          \\ \hline
            & EACC & 92.904 & \textbf{92.892}  & 92.814  & 92.362          & 92.302           & 91.841       & 92.719  &  90.954 & 92.890          & 92.781          \\
Densenet121 & EASR & /      & 99.724           & 98.558  & 99.780          & \textbf{100.000} & 94.375       & 99.943  &  11.046 & 99.015         & 98.884          \\
            & TASR & /      & \textbf{99.958}  & 94.001  & 97.966          & 97.831           & 94.447       & 99.896  & 46.032 & 99.948          & 99.859          \\ \hline
            & EACC & 91.196 & 91.966           & 91.934  & 92.270          & 92.524           & 92.119       & 92.312  & 91.713       & \textbf{92.370} & 92.152          \\
ResNet18    & EASR & /      & 99.760           & 99.126  & 99.904          & \textbf{100.000} & 93.766       & 98.916  & 14.971 & 99.990          & 99.801          \\
            & TASR & /      & \textbf{100.000} & 96.627  & 96.035          & \textbf{100.000} & 94.447       & 90.607  &  34.350 & 99.377          & 95.074         \\ \hline
            & EACC & 91.670 & 91.350           & 90.886  & \textbf{91.380} & 88.210           & 90.7900      & 91.190  & 90.750       & 90.851          & 90.307          \\
VGG11       & EASR & /      & \textbf{99.666}  & 99.734  & 97.852          & 21.714           & 93.480       & 99.861  & 10.288 & 99.162          & 90.737          \\
            & TASR & /      & \textbf{100.000} & 100.000 & 72.185          & 45.885           & 97.146       & 98.339  & 32.849 & 91.218         & 95.070          \\ \hline
\end{tabular}
\end{table*}

To showcase the universality of the proposed methodology, we utilized the Stochastic Gradient Descent (SGD) optimizer to train the model over a span of 100 epochs on four distinct model architectures namely VGG11 \cite{simonyan2014very}, ResNet18 \cite{he2016deep}, Densenet121 \cite{huang2017densely} and Alexnet \cite{krizhevsky2017imagenet}. During the training of ResNet18 and Densenet121, the optimizer was initialized with a learning rate of 0.1, while being subject to decay of 0.01 and 0.0001 at the 20th and 80th epochs, respectively. In addition, the optimization process for these models was performed with a momentum value of 0.9 and a weight decay of 0.0005. Conversely, for VGG11 and Alexnet, the optimizer was initialized with a learning rate of 0.01, which decayed to 0.001 and 0.00001 at the 20th and 80th epochs, respectively. The poisoning rate of 0.1 was applied to the training set. Notably, the final model performance was determined based on the result obtained from the last epoch of training. We repeated the training process for each model five times and calculated the average value as the final experimental result.

\begin{figure}[t]
  \centering
  \includegraphics[width=\linewidth, trim= 60 150 50 100,clip]{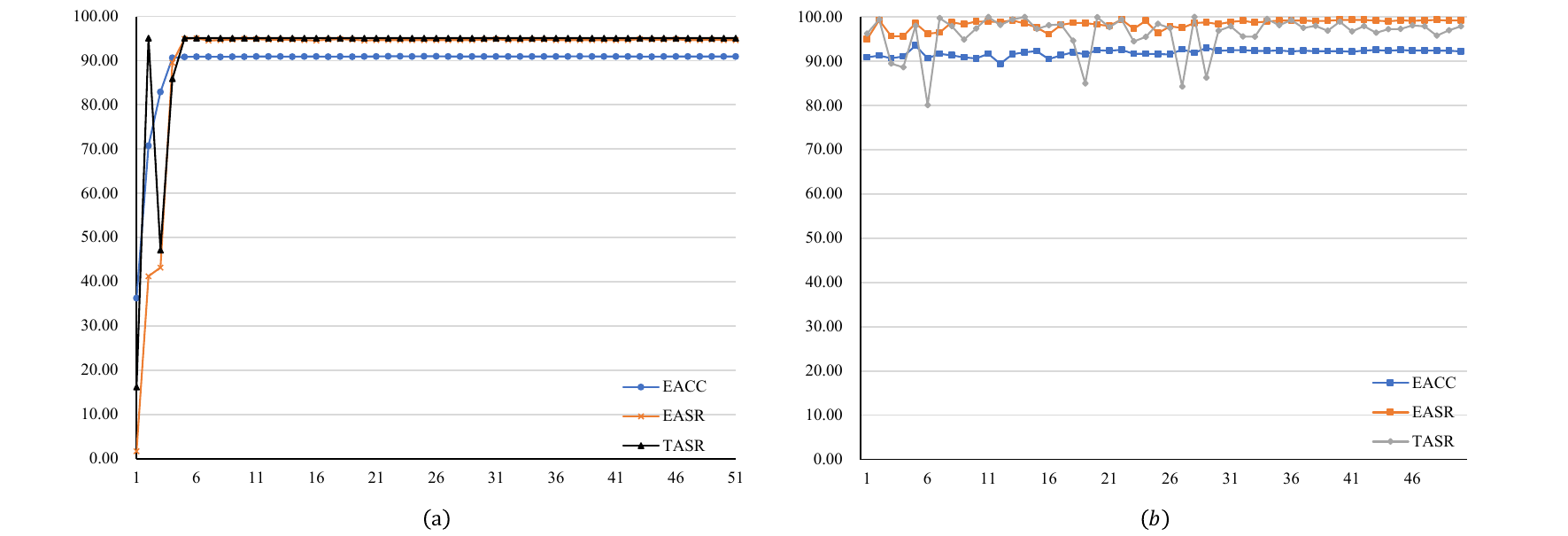}
  \caption{Performance of teacher and student models against fine-tuning defenses. (a): Student model. (b): Teacher model}\label{fig4.2.1}
\end{figure}

We evaluate the efficacy of the proposed method by comparing it with several classical backdoor attacks, namely Badnets \cite{gu2017badnets}, Blended \cite{liu2017neural}, Wanet \cite{nguyenwanet}, and IBSSA \cite{li2021invisible}, and report the comparative results in Table \ref{tab4.2.1}. In addition, we use PhysicalBA \cite{li2021backdoor} to physically simulate the above-mentioned attacks.
Specifically, PhysicalBA simulates the effect of shooting from multiple angles in a physical scene using data augmentation methods such as horizontal flipping, rotation, and color adjustment.
Figure \ref{fig4.2.1} depicts a comparison between our methodology and four distinct backdoor attacks in terms of invisibility. 
The triggers of Badnets and Blended are visible, and the poisonous image of ISSBA presents obvious colored patches. 
Figure \ref{fig4.2.1} shows that the triggers of Badnets, Blended, and ISSAB are still visible even with PhysicalBA. 
The trigger of Wanet is the most invisible, and the human vision system cannot distinguish the difference between the poisonous image and the original image.
Nevertheless, magnifying the residual image by a factor of ten reveals the clear appearance of the trigger.
% It is difficult to convince that the aforementioned backdoor can circumvent detection algorithms such as \cite{zeng2021rethinking}. 
Notably, the invisibility of our approach is perfect since no pixel in the original image is changed.

Considering the special properties of triggers in the proposed method, we define three performance evaluation indicators: benign Evaluation dataset ACCuracy (EACC), Validation dataset Attack Success Rate (EASR), and Test dataset Attack Success Rate (TASR).
It is worth noting that in the classical attack scheme, the benign verification dataset consists of 10,000 images. 
However, in our approach, we reduce the benign verification set to 9,000 images by excluding the 1,000 images that trigger the phone shooting. 
in calculating the attack success rate
Similarly, the classical attack introduces triggers to all images in the validation dataset, resulting in 10,000 poisonous data images. 
In contrast, our approach only involves 1,000 images taken by a specific trigger phone.
Additionally, the classical attacks employ 1,500 images to evaluate the TASR, while only 150 images are available for our approach.
% Although the total number of images between the different attack is unequal, we treat them as roughly equivalent since we present the final results in the form of percentages.

Overall, our method is comparable to classical backdoors in terms of attack performance. 
Specifically, Our EACC is close to the clean model under various model frameworks, and especially our EACC exceeds all other attacks under the Alexnet architecture. 
Our method is the only attack where EASR and TASR exceed 90$\%$ under each model architecture. The EASR of Badnets without PhysicalBA arrives at 89.015$\%$ on Alexnet. 
Similarly, the TASR Blended with PhysicalBA is 77.478$\%$ on Alexnet. Surprisingly, the ESAR of Wanet with PhysicalBA on VGG11 is only 10.288$\%$, which is equivalent to the backdoor not being implanted. and EASR of ISSBA without PhysicalBA on VGG11 is only 31.714$\%$.
The attack performance of the majority backdoor decreases within PhysicalBA, especially for Wanet. The EASR of Wanet with PhysicalBA decreases to about 10$\%$ and the EASR decreases to less than 40$\%$ under all architectures. Badnets and Blended also have varying degrees of attenuation. IBSSA is the least affected by PhysicalBA and even improves its performance relative to physical simulating. Under VGG11 and Alexnet architecture, the EASR and TASR of ISSBA improve significantly.

% TASR and EASR that are slightly lower than those of Badnet and PhysicalBA, but higher than other methods. However, the performance of Wanet is notably poor, with TASR and EASR decreasing to 70$\%$ and 60$\%$, respectively. On the Densenet121 architecture, all schemes perform well, with proposed method achieving slightly lower EACC than other attacks and second only to Badnet in TSAR. When tested on the Resnet18, ours achieves TASR almost 100$\%$, surpassing the invisible backdoor Blended, Wanet, and IBSSA. Our EACC is second only to Badnet but is higher than that of benign models. When evaluated on the VGG11 model, our EACC even exceeding that of the clean model. However, Wanet and ISSBA experience a considerable decline in TASR, with ISSBA dropping below 50$\%$. Regarding EASR, ISSBA drops to around 20$\%$ at one point.

\begin{figure}[t]
  \centering
  \includegraphics[width=\linewidth, trim= 65 115 65 140,clip]{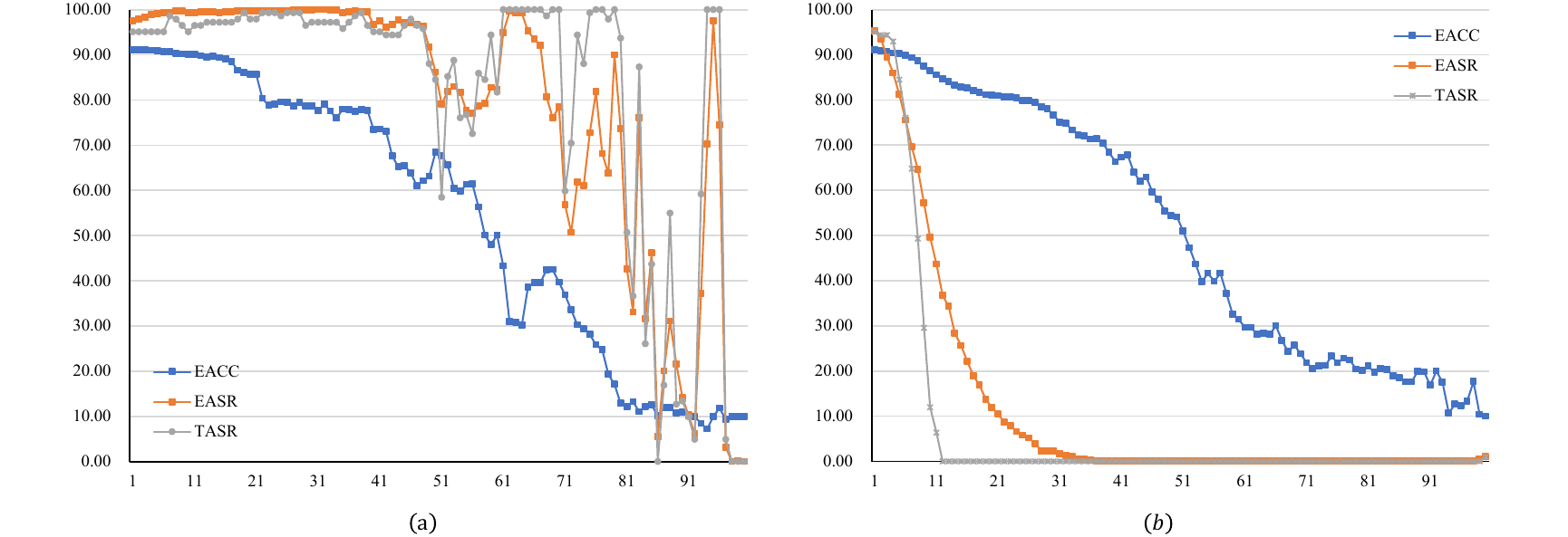}
  \caption{Performance of Badnets and our method against prnuing defenses. (a): Ours. (b): Badnets}
  \label{fig4.2.2}
\end{figure}

\subsection{Defense}

The invisibility of our approach is perfect since no pixel in the original image is changed. 
Thus we do not consider defense strategies regarding sample detection but focus on backdoor detection and excision of the model.
We select four defense strategies of fine-tuning \cite{liu2018fine}, pruning \cite{liu2018fine}, NAD \cite{li2021neural}, and NC \cite{wang2019neural} to evaluate the proposed method.

$\boldsymbol{Fine\ Tuning}$: We select the victim model with Resnet18 architecture for fine-tuning defense experiments. 
Fine-tuning model 50 epochs with optimizer SGD. The initial learning rate is 0.1 and decay to 0.01 at 3rd epoch.
Figure \ref{fig4.2.1} illustrates the trends of EACC, EASR, and TASR for teacher and student models as the number of fine-tuning epochs increases. 
Under the impact of a large learning rate in the early stage of the student model, EACC, EASR, and TASR all decline sharply. 
With the increasing of epochs, the three indicators gradually rise and stabilize at more than 90$\%$. 
The teacher model is unaffected nearly by fine-tuning, without the TASR slight fluctuations.

\begin{table}[t]
\caption{Performance comparison of various attack methods under NAD defense.}\label{tab4.3.1}
\begin{tabular}{ccccccc}
\hline
Attack &      & Badnets  & Blended & Wanet  & ISSBA  & Ours   \\ \hline
       & EACC & 90.750  & 91.190  & 90.160 & 91.900 & 90.807 \\
NAD    & EASR & 99.550  & 89.260  & 98.650 & 90.920 & 94.024 \\
       & TASR & 100.000 & 74.312  & 77.115 & 64.401 & 95.070 \\ \hline
\end{tabular}
\end{table}

$\boldsymbol{Puning}$: We respectively select the Resnet18 architecture victim model under Badnets and our attack for pruning defense experiments. We select "layer2" as the pruning layer, and 1$\%$ of neurons are pruned off each epoch. 
The experimental results are shown in Figure \ref{fig4.2.2}. 
With the increase of pruned neurons, the EACC of the two models gradually decreases.
The EASR and TASR of the proposed method remain at 95$\%$ before the 40th epoch. while the TASR of Badnets decreases sharply when pruning is carried out at 10 epochs, and the EAST drops to 0 in the 32nd epoch.

$\boldsymbol{NAD}$: We select the victim model under multiple attacks of Resnet18 architecture for NAD defense experiments, first fine-tuning for 20 epochs, and then distillation learning for 50 epochs. Considering that the victim must maintain the EACC of the model, we adaptively select the learning rate according to the standard that the EACC of models is not less than 90$\%$, rather than selecting a uniform learning rate. 
The experimental results are shown in Table \ref{tab4.3.1}. 
The EASR and TASR of our method are barely affected by NAD, EASR decreases 5$\%$ and TSAR remains unchanged. 
Blended, ISSBA, and Wanet TASR all decline remarkably, attenuating to less than 80$\%$. 
To sum up, the proposed method has no disadvantage compared with the classical backdoor attacks against NAD.

\begin{figure}[t]
  \centering
  \includegraphics[width=1.0\linewidth, trim= 110 160 90 150, clip]{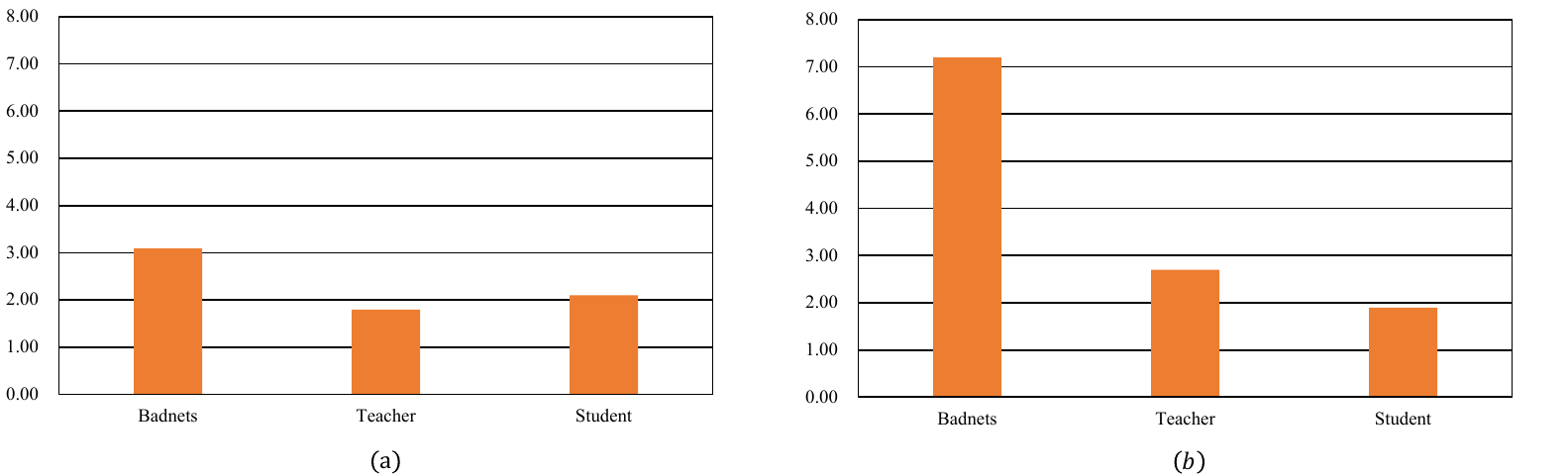}
  \caption{Performance comparison of our method and Badnets against NC detection with different architecture.(a) ResNet18, (b)VGG11}
  \label{fig4.2.3}
%   \Description{A woman and a girl in white dresses sit in an open car.}
\end{figure}

$\boldsymbol{NC}$: To verify the imperceptibility of the proposed method, we select the most classical NC detection algorithm to detect whether the model injected a backdoor or not. 
According to the classical assumption, the defender uses the labeled 0.05$\%$ data set to detect the anomaly of the model. 
The model is considered to have been implanted with a backdoor when the outlier value is greater than 2.5. 
We select the models of two model architectures of Resnet18 and VGG11 under Badnets and proposed an attack to conduct NC detection experiments. 
The experimental results are shown in Figure \ref{fig4.2.3}, Badnets fail NC detection under both architectures, and the outlier value of the victim model under VGG architecture exceeds 7.0. 
Under the Resnet18 architecture, the teacher model in the proposed method has an outlier value of 2.1, which is less than the threshold value of 2.5, and successfully avoids NC detection. 
However, the teacher model under the VGG11 architecture, with an outlier value of 2.7, exceeds the threshold and fails to pass the detection. 
Under both architectures, the student model has lower outliers than the teacher model, and all of them pass the detection. 
It indicates that our attack can bypass the NC.
% This may be due to the backdoor feature and image semantic image fused during the feature distillation stage.

\section{CONCLUSION}

In this paper, we present a pioneering approach to clandestinely injecting backdoors into classical models without altering pixels. Our innovative method harnesses camera fingerprint features as triggers, ensuring exceptional invisibility. Moreover, our approach tackles the challenge of physically implementing a backdoor attack. Specifically, when feeding images captured by a specific camera into the model, they induce the desired output for adversaries, while images taken by other cameras are accurately identified. To assess the effectiveness of our approach, we carefully curated a dataset comprising 21,500 images captured by various mobile phones. The experimental results demonstrate that our method can successfully attack diverse models with different architectures. It maintains a comparable attack success rate, benign sample accuracy, and anti-defense robustness against classical attacks.
 
\begin{acks}
This work was supported by the National Natural Science Foundation of China (U20B2051, U1936214, U22B2047).
\end{acks}

\bibliographystyle{ACM-Reference-Format}
\bibliography{refs}

% %%
% %% If your work has an appendix, this is the place to put it.
% \appendix









\end{document}